# ReenactGAN: Learning to Reenact Faces via Boundary Transfer

Wayne Wu[1]*, Yunxuan Zhang[1]*, Cheng Li[1], Chen Qian[1], Chen Change Loy[2]

[1] SenseTime Research
[2] Nanyang Technological University
{wuwenyan,zhangyunxuan,chengli,qianchen}@sensetime.com  ccloy@ntu.edu.sg

**Abstract.** We present a novel learning-based framework for face reenactment. The proposed method, known as ReenactGAN, is capable of transferring facial movements and expressions from an arbitrary person's monocular video input to a target person's video. Instead of performing a direct transfer in the pixel space, which could result in structural artifacts, we first map the source face onto a boundary latent space. A transformer is subsequently used to adapt the source face's boundary to the target's boundary. Finally, a target-specific decoder is used to generate the reenacted target face. Thanks to the effective and reliable boundary-based transfer, our method can perform photo-realistic face reenactment. In addition, ReenactGAN is appealing in that the whole reenactment process is purely feed-forward, and thus the reenactment process can run in real-time (30 FPS on one GTX 1080 GPU). Dataset and model are publicly available on our project page[†].

## 1 Introduction

Face reenactment aims at transferring one source person's facial expression and movements to another target person's face. Faithful photo-realistic facial reenactment finds a wide range of applications, including film production, video conferencing, and augmented reality (*e.g.*, virtual YouTuber). Thanks to the increasingly accurate and reliable 3D facial model fitting [2, 9] and landmarks detection [6, 19, 39, 50–52, 60, 61] techniques on RGB-D and RGB cameras in recent years, many impressive face reenactment methods are proposed [41, 35, 34, 32, 3, 18]. Most existing approaches represent face as a predefined parametric 3D model. These methods typically involve tracking and optimization to fit a source video into a restrictive set of facial poses and expression parametric space, and then render the manipulated target output. In general, optimization-based methods can handle background regions better, compared to feed-forward based method that can only support pixels generation around the face. Nevertheless, a pre-defined parametric 3D model can hardly capture all subtle movements of the human face. In addition, these works require large efforts and delicacy designs of complex parametric fitting. Considering the algorithmic complexity of these approaches, few of them are open-sourced.

---

*Equal contribution. This work was done during an internship at SenseTime Research.

[†]Project Page: https://wywu.github.io/projects/ReenactGAN/ReenactGAN.html

The emergence of Generative Adversarial Network (GAN) based approaches, *e.g.*, Pix2Pix [15] and CycleGAN [57], offers an appealing and succinct alternative for face reenactment. Nonetheless, despite the success of GAN in many image-to-image transfer applications [27, 15, 57], training a pure learning-based method for face reenactment is *non-trivial*: (1) Face images are captured under very different poses, expressions and lighting conditions. Thus learning a direct face-to-face mapping based on limited samples but covering all variances is hard. A conventional state-of-the-art GAN such as CycleGAN would generate unnatural images in extreme conditions such as large pose, or fail on unseen images, as shown in Figure 1. (2) No pairwise data is available as we can hardly match diverse expressions given an arbitrary set of source and target videos. (3) We wish to perform *many-to-one* mapping, *i.e.*, reenact a specific target with just a single model given any source faces. This scenario violates the assumption of CycleGAN since an inverse mapping (one-to-many) does not exist.

To address the first two challenges, we are required to define a space or medium that allows an effective and robust transfer of facial movement and expressions. Inspired by previous facial model fitting approaches, we propose the use of facial contours or boundaries as a compact medium to capture facial geometric variances. Specifically, we map a source face into a latent boundary space, which we wish facial movement and expressions are faithfully kept. We then adapt the space to the specific target person and decode for the appearance.

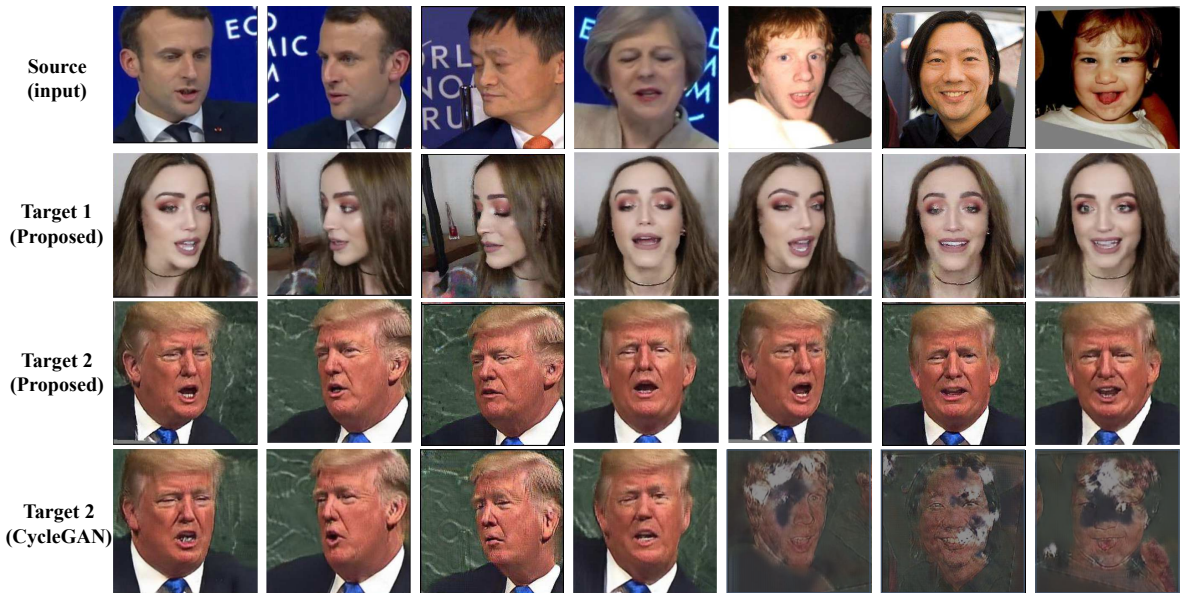

**Fig. 1.** The proposed ReenactGAN is capable of manipulating a target face in a video by transferring movements and facial expressions from an arbitrary person's video. In contrast to CycleGAN [57], ReenactGAN can comfortably support the reenactment of large facial movements. CycleGAN is infeasible to transfer unseen data as shown in the last three columns.

Introducing facial boundaries as the latent space is beneficial for face reenactment. *Firstly*, the mapping between face images and the facial boundary space relates to the well-studied face alignment problem. Thanks to contemporary facial landmark detection methods [61, 50, 19, 6], face boundaries can be obtained accurately and robustly under large poses, diverse expressions and extreme lighting conditions. This is a unique advantage that is not available in direct raw pixels-based mapping. *Secondly*, transferring diverse expression with unpaired data is simplified. The availability of large-scale face alignment training sets provides high-quality paired training data that consist face images and the corresponding boundaries. Learning without paired data only happens in the well-defined boundary space, whilst the input face encoding and target face decoding processes can fully leverage the power of pairwise face alignment data.

Our final challenge is to resolve the *many-to-one* mapping problem. We address this problem by formulating a GAN-based transformer to adapt the boundary space of input face to the target face. To ensure the quality of transformation, we constrain the process with a cycle loss, an adversarial loss, and a shape loss in the PCA space. With the target-specific transformer, we can reenact a target face based on images or videos from arbitrary sources. Overall, the proposed **ReenactGAN** hinges on three components: (1) an encoder to encode an input face into a latent boundary space, (2) a target-specific transformer to adapt an arbitrary source boundary space to that of a specific target, and (3) a target-specific decoder, which decodes the latent space to the target face. ReenactGAN is easy to re-implement and distribute since each component in the framework is a feed-forward network, and the only training material is source-target videos and a facial alignment training set.

We summarize our contributions as follows:

- We introduce the notion of 'boundary latent space' for face reenactment. We found that facial boundaries hold sufficient geometric information to reenact a face with rich expressions but being relatively 'identity-agnostic' in comparison to direct mapping with raw pixels. Importantly, the boundary space is more robust to challenging poses, expressions and lighting conditions.

- Based on the notion of boundary latent space, we propose a novel learning-based face reenactment framework. All components are feed-forward. In contrast to traditional model-based methods, the proposed ReenactGAN is easier to train and implement.

- We introduce target-specific transformers in the latent space to achieve many-to-one face reenactment, which is otherwise impossible using a conventional GAN-based image-to-image transfer method.

## 2   Related Work

**Face Reenactment:** Most of the existing studies can be categorized as a 'model-based' approach. These methods typically consist of three steps: (1) Face capturing, *e.g.* tracking face templates [41], using optical flow as appearance and velocity measurements to match the face in the database [22], or employing either RGB [4] or RGB-D camera [34] to capture face movements. Recent advances of facial landmark detection methods [51, 5] enable us to effectively track input's facial component like eyes and mouth. (2) Once the facial movement is captured, many studies will subsequently to fit the movement in a parametric space or model, including head pose [41, 44], eye gaze [36, 44], or PCA coefficients over 3D model bases [35] and even detail 3D face meshes [48]. (3) Once a model is fitted, the next step is to re-render a new video. Garrido *et al.* [8] directly retrieves similar pose to the source from the target video, and render new video through morphing. A similar strategy is employed in [35] to optimize the inner-mouth generation process. Retrieval-based methods are arguably low in computational efficiency [35, 40]. Recently, CycleGAN provides a new feasible solution for face reenactment. To our best knowledge, although there are no published peer-reviewed papers in the literature, some interesting work was released or demonstrated in the community [53, 16, 47]. In contrast to the CycleGAN that can only handle a single source person and one target, our framework aims at solving the harder many-to-one problem, which permits more practical usages.

**Generative Adversarial Network (GAN) and CycleGAN:** Generative Adversarial Networks (GAN) [10] has been extensively used and extended for image generation including facial images. By employing the adversarial loss, one can map a low-dimensional noise vector input [10], a fashion design coding [59], an unrealistic rendered facial image [28] or a text description [54] to a realistic image. Zhu *et al.* [57] has shown that by adding a cycle consistency loss, CycleGAN achieves impressive results in learning the transformation function between two domains. Recently, Mueller *et al.* [27] leveraged a geometric consistency loss to preserve hand pose based on CycleGAN. Our method is different in directly taking image-to-image translation on the geometric latent space rather than adding an auxiliary geometric constraint. Xu *et al.* [53] applied CycleGAN to learn the transformation between a specific person pair. They added specific discriminators on different facial components to enhance the performance of each local part. In contrast to [53], our ReenactGAN first maps all faces into a boundary latent space and then decodes it to each specific person. With the proposed target-specific transformer, each decoder can reenact arbitrary person to a specific target based on the adapted boundary space, thus achieving many-to-one reenactment efficiently and conveniently. Introducing the boundary space also improves facial action consistency and the robustness for extreme poses.

## 3   Face Reenactment via Boundary Transfer

The proposed framework, ReenactGAN, is depicted in Figure 2. ReenactGAN can be divided into three components: a boundary encoder, a target-specific many-to-one transformer, and a target-specific decoder. Each component is a feed-forward network. In the test phase, a query face will be forwardly passed through each component. The remainder of this section will be organized as follows: Sec. 3.1 presents the encoder, decoder, and the joint reconstruction loss; Sec. 3.2 describes the boundary latent layer; Sec 3.3 explains the details of the target-specific many-to-one transformer.

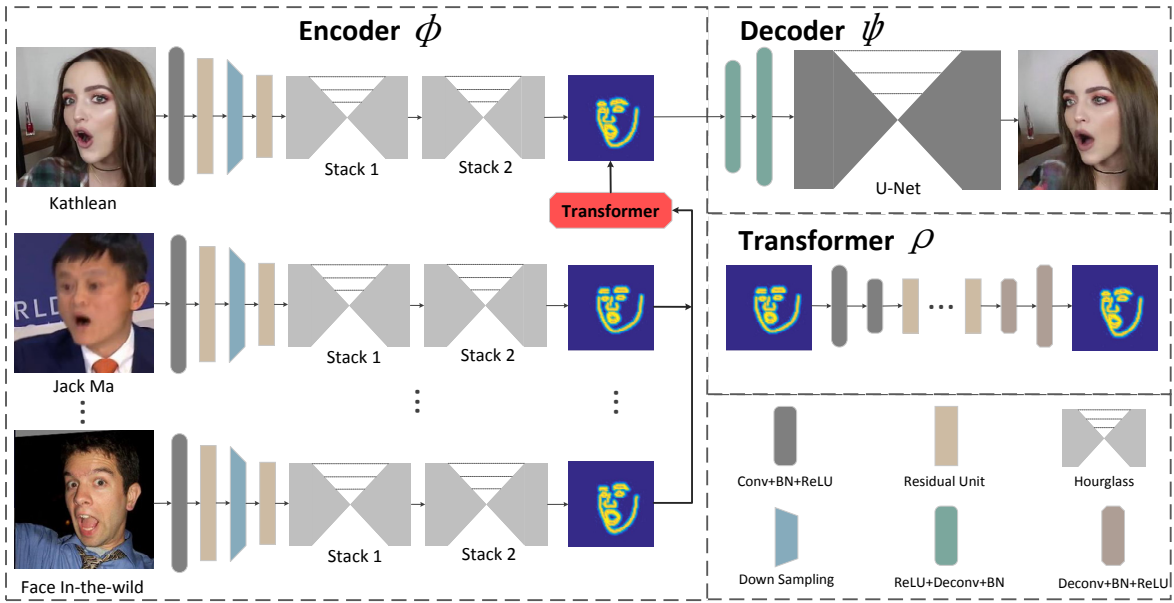

**Fig. 2. The framework of ReenactGAN:** There are three main components, all of which are feed-forward networks. In the test phase, an encoder $\phi$ encodes an image $x$ into the boundary space $b$. A transformer $\rho$ adapts an arbitrary face's boundary into the target's. A decoder $\psi$ decodes the adapted boundary $\rho(b)$ to a target's face $t$.

### 3.1 Encoder and Decoder

Our goal is to find a function pair: an encoder $\phi : \mathcal{X} \to \mathcal{B}$ and a target-specific decoder $\psi_{\mathcal{T}} : \mathcal{B} \to \mathcal{X}$, where $\mathcal{X}$ represents faces at the raw pixel space, and $\mathcal{B}$ denotes the boundary space. The encoder maps a face $x \in \mathcal{X}$ into a latent space $b \in \mathcal{B}$. The target-specific decoder decodes a latent boundary $b$ back to a specific person's face $t \in \mathcal{T} \subset \mathcal{X}$. We design our encoder and decoder following the state-of-the-art Pix2Pix approach [15] to synthesize photo-realistic faces. Specifically, we adopt a combined loss widely used in generation tasks [30, 21, 45]:

$$L(\psi \cdot \phi, \theta) = L_{\text{GAN}}(\psi \cdot \phi, \theta) + L_{\ell_1}(\psi \cdot \phi) + L_{\text{feat}}(\psi \cdot \phi). \quad (1)$$

Without the loss of generality, we denote $\psi_{\mathcal{T}}$ as $\psi$. The first term $L_{\text{GAN}}$ is an adversarial loss, which employs a discriminator $\theta$ to distinguish the real sample $t$ and the reconstructed sample $\psi \cdot \phi(t)$. The second term $L_{\ell_1}$ is an L1 reconstruction loss. The third term $L_{\text{feat}}$ measures the L2 distance between relu2_2 and relu3_3 features of the VGG-16 [31] network. The combination of these three losses is widely used for image reconstruction to generate sharp and realistic outputs. It is noteworthy that our target-specific decoder $\psi_\mathcal{T}$ does not forcefully decode $\mathcal{B}$ back to the full face space $\mathcal{X}$. Instead, each target-specific decoder only focuses on one target person's face subset $\mathcal{T}$ such that high-quality specific-person synthesis can be achieved. This notion is shown in Figure 3. To achieve this goal, we collect a set of target's faces $\{t_1,\ldots,t_N\} \in \mathcal{T}$ during the training stage to train the $\psi_T$. The encoder $\phi$ is shared and all $\{\psi.\}$ and $\phi$ are jointly trained.

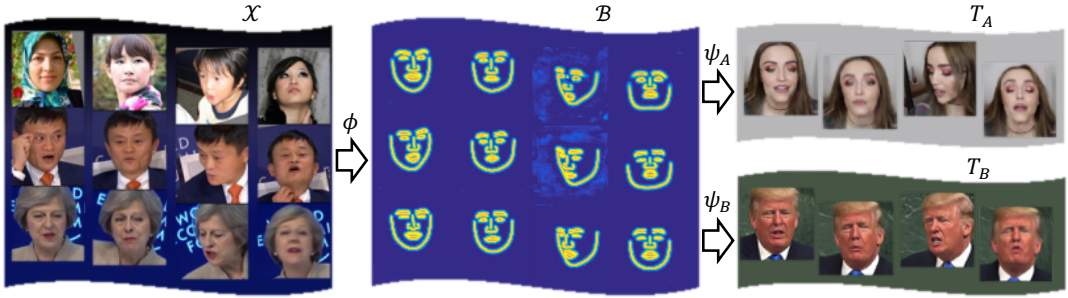

**Fig. 3. An illustration of encoding and decoding process:** We first encode all faces into a latent boundary space $\mathcal{B}$. Then for each target $\mathcal{T}$, we have a specific-decoder $\psi_\mathcal{T}$ to decode latent boundary to the corresponding face $t \in \mathcal{T} \subset \mathcal{B}$.

The loss in Eqn.(1) effectively train the pair of $\psi, \phi$, which allows us to map face images into the latent space $\mathcal{B}$, and vice versa. Our next challenge is to bridge the gap between the source boundary space and the target boundary space. We discuss the challenges and our solution in the next section.

### 3.2   Boundary Latent Space

Recall that Eqn. (1) only facilitates the mapping between the pixel space $\mathcal{X}$ and latent space $\mathcal{B}$. In this section, we discuss how to design the latent space $\mathcal{B}$. We hope that the latent space could satisfy two properties. Firstly, $\mathcal{B}$ should be sensitive to facial expression but less so on identity. In particular, assuming two faces $x_1, x_2$ with different identities but the same expression, they should be mapped to a nearby location in the the boundary space $\mathcal{B}$. Secondly, $\mathcal{B}$ should encompass rich structural information to support the decoding process for appearance.

To this end, we design the latent space as a stack of $K$ boundary heatmaps, $\{M_i\}_{i=1}^K$, each of which represents the contour of a specific facial part, *e.g.*, upper left eyelid and nose bridge. Each heatmap maintains sufficient spatial resolution of $64 \times 64$ to retain the structural information of a face. Some examples of heatmaps (projected to a single map for visualization) are shown in Figure 3. As can be observed, the boundary heatmaps are appealing in that they are not affected by background clutter, lighting, and facial textures. In comparison to the raw pixels space, faces with the same expression are naturally closer to each other in this boundary space.

To constrain the learning of this latent space, during the training of encoder $\phi$ and decoder $\psi$, we add a L1 loss between the encoder output, $\phi(x)$, and its associated ground-truth boundary heatmaps $\{M_i\}$. We prepare the ground-truth boundary heatmaps by exploiting existing facial landmark datasets [50, 20]. Specifically, each face image is defined with a set of $P$ landmarks represented as $\mathcal{S}$. The subset of landmarks corresponding to the $i$-th facial part, denoted by $S_i \subset \mathcal{S}$, is firstly interpolated to get a dense boundary line. We then set points on the line to 1 and others as 0. Note that this form of sparse binary heatmap would cause problem since regressing to a single value point at the boundary is difficult and highly nonlinear. We address this problem by applying a Gaussian smooth on the boundary so that regression can be done on a confidence map located in the immediate vicinity of the boundary location.

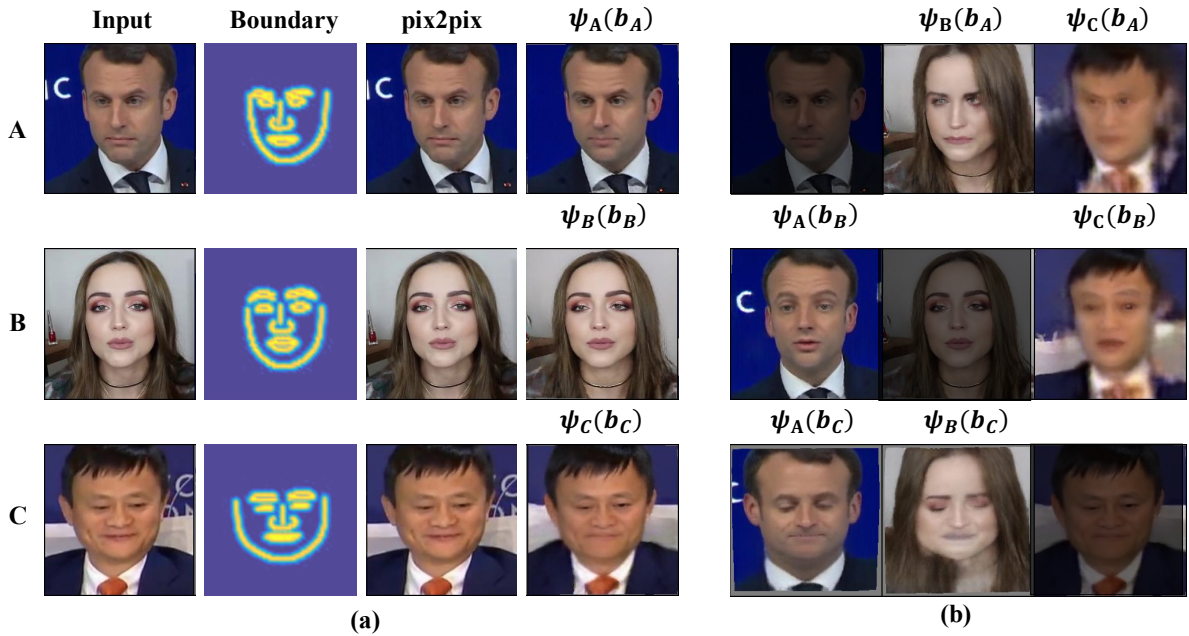

**Fig. 4.** (a) **Face reconstruction with and without boundary latent space**. From left to right: 1. Raw image input; 2. Boundary map used for latent supervision; 3. Pix2Pix's reconstruction results without boundary latent space; 4. Reconstruction results with boundary latent space. (b) **Face reenactment without boundary transformer**. We apply a target-specific decoder using other's boundary as input. Here no adaptation is made on the boundary for a specific target, therefore failures can be seen if face boundaries do not match well with the decoder.

In Figure 4(a), we compare the reconstruction quality of faces with and without using the boundary latent space. Reconstruction not using the latent space is equivalent to Pix2Pix [15]. As can be seen from the third and fourth columns, introducing boundary latent space do not affect the reconstruction quality, justifying the stability of the learned boundary space. Figure 4(b) shows some preliminary results of applying target-specific decoder, such as $\psi$ of Emmanuel Macron (A), with boundary heatmaps of Jack Ma (C) or Kathleen (B) as input (see the last row). When two faces share a similar shape, *e.g.*, persons A and B, the reenactment is satisfactory. However, when the face shape differs, a decoder would suffer in decoding a mismatched boundary input causing artifacts

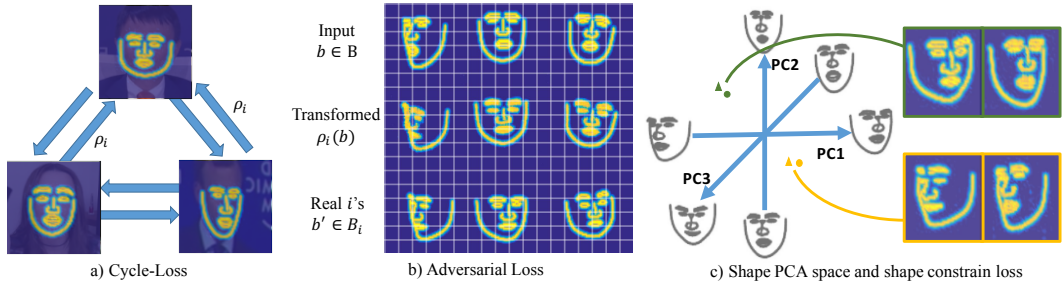

a) Cycle-Loss         b) Adversarial Loss         c) Shape PCA space and shape constrain loss

**Fig. 5. Transformer's loss:** (a) For each person we maintain a transformer. We enforce all transformers to satisfy the rule of cyclic consistency. (b) For each person we also have a discriminator $C_i$ to predict if a boundary map belongs to him/her. c) We constrain the input and output of each transformer to be close in PCA subspace.

in its generated image. The observation motivates the need of the boundary transformer, which we will discuss in the next section.

We wish to point out that facial boundaries are by no means the only medium to serve as the latent space. We believe any medium that could faithfully represent facial expression and movement, yet near-identity-agnostic can be used here. Facial boundaries are chosen here as we have access to public large-scale landmark dataset for constraining the learning of the space. Other medium such as facial expression space and densely annotated coordinates [12, 11] is applicable if large datasets are available to constrain the learning of the space.

### 3.3  Boundary Transformer

As shown in Figure 4(b), applying a target-specific decoder on the boundary heatmaps of other person may lead to severe artifacts when there is a large structural gap between the face shapes of the source and target. We address this problem through target-specific transformers, $\rho_\mathcal{T}$. The design of $\rho_\mathcal{T}$ aims to bridge the gap between an arbitrary people's and the target's boundary space, so as to transfer an arbitrary boundary into a specific target. Formally, it maps $\phi(\mathcal{X})$ to $\phi(\mathcal{T})$.

The learning of this transformer can be formulated in the CycleGAN framework [57] since $\phi(\mathcal{X})$ and $\phi(\mathcal{T})$ are unpaired (they are obtained from videos of different persons). Nevertheless, as shown in Figure 3, $\mathcal{T}$ is only a subset of $\mathcal{X}$, finding the inverse mapping from $\phi(\mathcal{T})$ to $\phi(\mathcal{X})$ is problematic since the mapping from an arbitrary people to one may exist, but the inverse is a multivalued function that cannot be easily solved by a feed-forward network.

Recall that $\rho_\mathcal{T}$ reduce the full set $\mathcal{B}$ to a subset $\mathcal{B}_\mathcal{T}$. If we only consider one $\rho_\mathcal{T}$, its inverse is multi-valued, and it is infeasible to establish the cycle consistency. However, if we consider multiple target-transformers at the same time and modify the cyclic loss of [57], we can circumvent the aforementioned issue. Let's assume that we have several training data of several targets, $\{\mathcal{T}_1, \ldots, \mathcal{T}_N\}$, where $\mathcal{T}_i \subset \mathcal{X}$. For brevity, we denote the image of one target's face in the boundary space $\phi(\mathcal{T}_i)$ as $\mathcal{B}_i \subset \mathcal{B}$. We hope to train a series of transformers, which can transform $\mathcal{B}$ to each $\mathcal{B}_i$. As illustrated in Figure 5, the loss of for training the transformers consists of three terms:

$$L(\{\rho.\}, \{C.\}) = L_{\text{cycle}} + L_{\text{GAN}} + L_{\text{shape}}, \quad (2)$$

where $\{\rho.\}$ is the set of transformers and $\{C.\}$ is the set of corresponding discriminators in the adversarial learning notion.

The first term is a loss that constrains the cycle consistency. Specifically, we let $b_i \in \mathcal{B}_i$ and we sample an arbitrary transformer $\rho_j$, where $j \neq i$. With this term we wish to define a loss such that $\rho_j(b_i)$ belongs to $\mathcal{B}_j$ and $\rho_i \cdot \rho_j(b_i)$ equals to $b_i$. The loss can be written as

$$L_{\text{cycle}} = \mathrm{E}_{i \neq j} \left[ \| \rho_i \cdot \rho_j(b_i) - b_i \| \right]. \quad (3)$$

The second term defines the vanilla GAN loss. We train a discriminator $C_i$ for each target person. The responsibility of a discriminator is to distinguish one target's real pose from transformed pseudo poses. The loss is defined as

$$L_{\text{GAN}}(\{\rho.\}, \{C.\}) = \sum_i \left( \mathrm{E}_{b_i \in \mathcal{B}_i} \log C_i(b_i) + \mathrm{E}_{b' \in \mathcal{B}} \log(1 - C_i \cdot \rho_i(b)) \right). \quad (4)$$

The third term is a shape constrain loss that encourages a transformed boundary to better follow its source. The loss is defined between the input and output of a transformer. Specifically, we first use a fully-connected layer to map the shape latent $b$ to a vector. We then compress the vector via PCA and only keep the first $M$ coefficients that capture the rough motion of the head. We denote this linear process as the function $R$ and define the shape constraint loss as

$$L_{\text{shape}} = \mathrm{E}_{b \in \mathcal{B}, i \in 1, \ldots, N} \left[ R(b) - R \cdot \rho_i(b) \right]. \quad (5)$$

## 4 Experiments

We evaluate face reenactment from two aspects: (1) Image quality – after we show the qualitative result in Sec. 4.1, we report a user study in Sec. 4.2. (2) Facial action consistency – to measure if the generated output correctly captures the expressions in the input face, we compare the response of facial action units [7, 25] in Sec. 4.3. Sec. 4.4 finally provides an ablation study on the losses of transformer.

**Celebrity Video Dataset:** For $\{\mathcal{T}_1, \ldots, \mathcal{T}_N\}$, we collect five celebrities' videos from YouTube, namely Donald Trump, Emmanuel Macron, Theresa May, Jack Ma and Kathleen. The average length is 30 minutes. These celebrities have quite different facial characteristics thus the videos are well-suited for robustness evaluation. All of the 200K faces are annotated with 98-landmarks using a semi-automatic methodology, in which each face is annotated by a state-of-the-art facial landmark detection method [50] followed by additional manual correction. The Celebrity Video Dataset (CelebV) are available on our project page.

**Boundary Estimation Dataset:** We combine two face alignment datasets to generate the ground truth of boundary heatmaps. The first one is WFLW [50] dataset, which provides 10000 photos with 98 annotated landmarks on each face. WFLW dataset enables us to generate 14 contours including nose bridge, right upper eyebrow, left upper eyebrow, *etc*. The second dataset is Helen [20], we use the annotation protocol of Helen [20] for the nose boundary. These two types of protocol are fused and then used to generate the ground truth of boundary heatmaps as discussed in Sec. 3.2. Finally, 15 ground truth boundary heatmaps are obtained for the training of Encoder.

**Pre-processing:** The faces were detected by Faster R-CNN [29], tracked by KCF tracker [13]. Faces belong to irrelevant people are removed by DeepFace [33] from the training set. Each face was normalized to a mean shape with rigid transformation and cropped to 256×256.

**Training Details:** Encoder is trained on 7500 faces of WFLW Dataset and then fine-tuned on CelebV Dataset, while Transformer and Decoder are both trained only on CelebV Dataset. Due to the limited space, we report the network architecture and training details in our supplementary material.

### 4.1   Qualitative Comparisons with State-of-the-arts

We compare ReenactGAN with two state-of-the-art methods. (1) CycleGAN: since there is no commonly accepted CycleGAN-based implementation for face reenactment, we implement our own based on [57]. (2) Face2Face [35]: Face2Face is not open-sourced. Comparisons are thus limited by our attempts to crop images/videos from its released YouTube demo.

Figure 6 compares our method with the two state-of-the-art methods. All three methods work well for frontal faces. By contrast, the proposed method works well on profile faces (Figure 6(a)), while being more effective in transferring expressions of unseen people to the target (Figure 6(b)). It is observed that the vanilla CycleGAN performs poorly on hair generation. We believe that this issue may be solved by increasing the power of discriminator [53]. Our method does not suffer from the same problem, by encoding into the boundary latent space, the hair information will be first excluded, then be rendered by the decoder. On the other hand, in contrast to Face2Face that only alters the appearance of inner-face, our model can additionally track the global head pose from the source video (Figure 6(c)).

### 4.2   A User Study on Face Reenactment Quality

We are curious about several questions: does the image has a sound quality, sharp and clear? Does the image contain strange artificial textures? Does the expression of the image look real? We are also interested in another question: during the training of CycleGAN [57], the model has already 'seen' some data. How good does it work on the unseen wild data?

To answer these questions, rather than taking general image quality assessment methods [46, 23, 24] or perceptual loss [58] for evaluation, we perform a user study since human observation is more direct and reasonable for the validation of perceptual realism. We ask 30 volunteers to compare the quality between

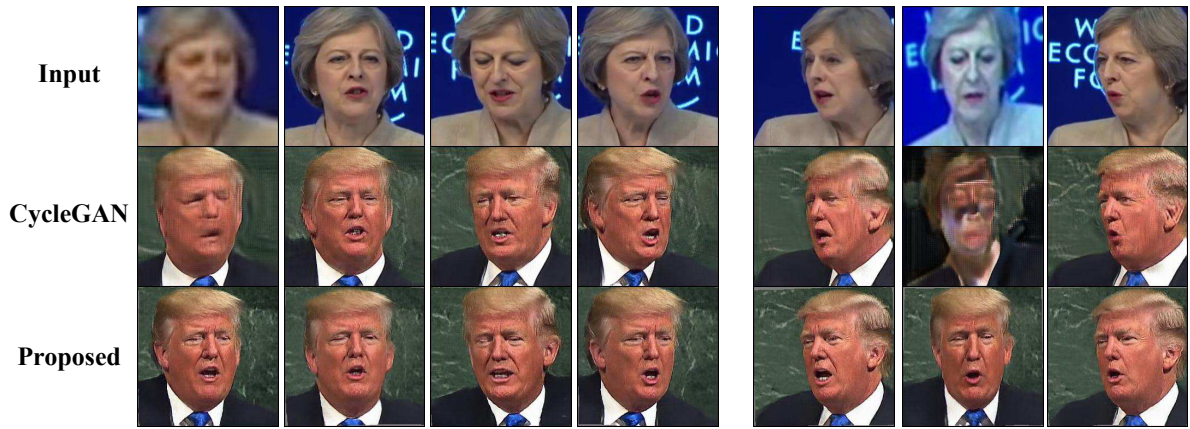

**(a) Frontal Input vs. Extreme Pose Input**

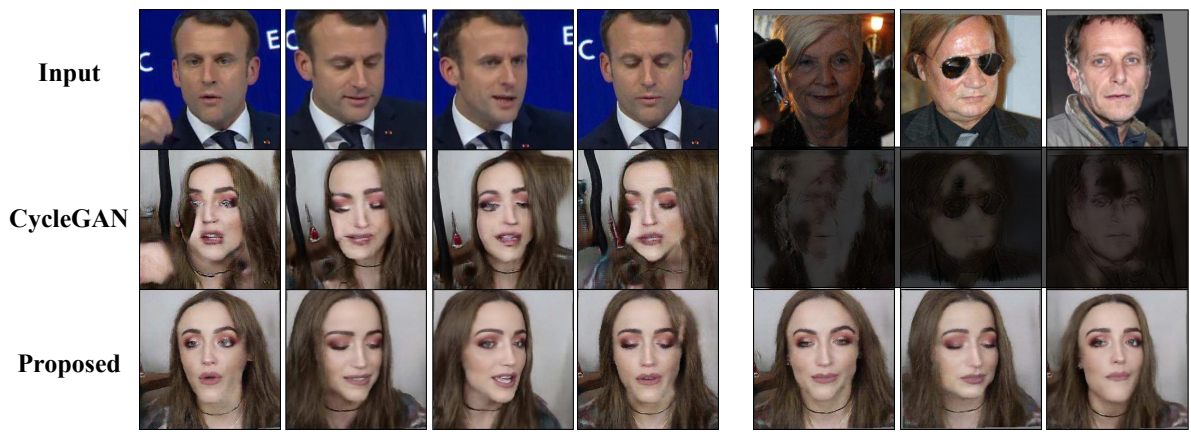

**(b) Test on Seen Data vs. Wild Input**

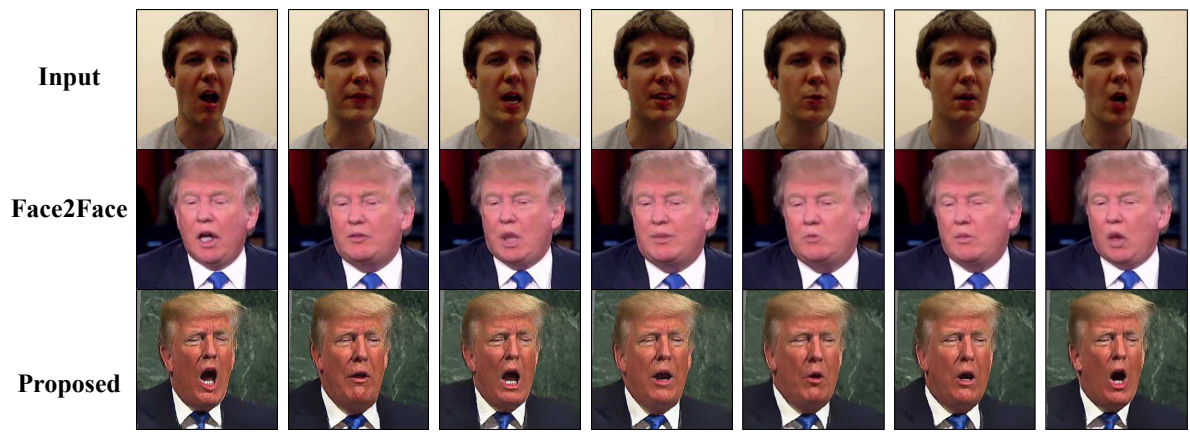

**(c) Compare with Face2Face**

**Fig. 6. Qualitative comparison with state-of-the-arts:** We compare ReenactGAN with a GAN-based method - CycleGAN[57] and model-based approach Face2Face [35]. In the second case we tried wild inputs that a model has never seen. Vanilla CycleGAN fails because it only learns the transformation between one source person and one target.

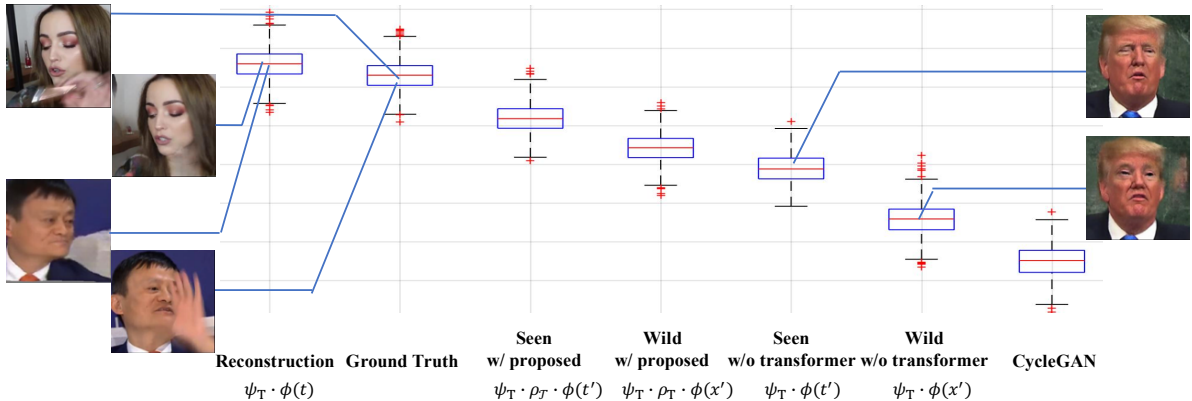

**Fig. 7. User study:** We compare the quality among the real video and 6 different groups of generated result. Then we computing their quality score via TrueSkill and visualize the statistic.

two generated photos. Specifically, we follow the protocol presented in [15, 55], in which each comparison is limited to 10 seconds. After numerous comparisons, we determine the score of each faces via TrueSkill algorithm [14] (an offline generalization of Elo-Rating). To avoid biases caused by photo contents (*e.g.*, one may perceive a photo that is more attractive as to have a better quality or may consider a frontal face as lower quality), we prepare our comparisons to have the same pose and similar expression.

We measure the quality among seven different settings: (1) Ground Truth, a.k.a $t \in \mathcal{T}$; (2) Reconstruction result of the ground truth, $\psi_\mathcal{T} \cdot \phi(t)$; (3) Seen data with ReenactGAN, $\psi_\mathcal{T} \cdot \rho_\mathcal{T} \cdot \phi(t')$, which $t' \in \mathcal{T}' \neq \mathcal{T}$; (4) Unseen data with ReenactGAN, $\psi_\mathcal{T} \cdot \rho_\mathcal{T} \cdot \phi(x')$, which $x' \in \mathcal{X}_{Test}$, also $x'$ does not belong to any trained target identity; (5) Seen data without transformer, $\psi_\mathcal{T} \cdot \phi(t')$; (6) Unseen data without transformer $\psi_\mathcal{T} \cdot \phi(x')$; (7) Seen data with CycleGAN.

We prepared 180 groups of results for this user study. For each group, we first picked a face in the target set. We then retrieved the most similar pose in-the-wild from both input set and source training set, and generated six results using different models. In each group, the full $C_7^2 = 21$ comparisons were evaluated. Then we used TrueSkill [14] algorithm to compute the score of each image. It is evident from Figure 7 that the proposed method outperforms CycleGAN in this user study. The transformer plays a crucial role in the reenactment. Unseen wild data will cause a performance drop but ReenactGAN can handle this challenge well. Interestingly, we found that $\psi_\mathcal{T} \cdot \phi$ would inpaint the occluded facial part, and thus volunteers sometimes marked the reconstructed result to have a better quality over the ground truth.

### 4.3   Facial Action Consistency

Besides image quality, we propose an interesting way to measure the effectiveness of face reenactment methods on transferring facial expressions. We borrow the notion of facial action units (AU) [17, 1, 37, 42, 56, 43, 38, 49]. Specifically, we train an action unit detector on DISFA [26] dataset, which provides 12 kinds of facial action units. We use 98 facial landmarks as input to train a MLP for each action unit. The reason why we only use facial landmarks as input is to prevent overfitting on this relatively small dataset. The 5-fold cross-validation accuracy is averagely 55.2% on DISFA [26], which is competitive with the state-of-the-art method (58.4% [49], the random guess is 8.3%).

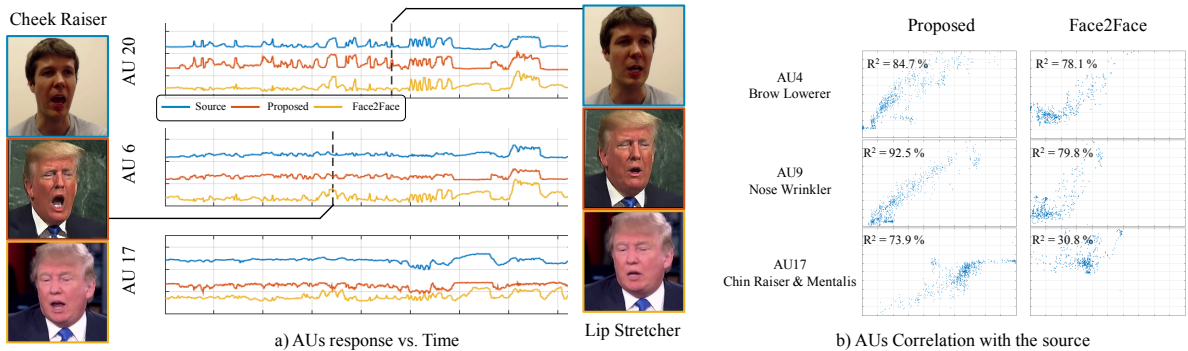

**Fig. 8. Facial action consistency:** (a) We use a facial action detector to obtain responses from ReenactGAN and Face2Face [35]. Our method is more accurate in transferring the contour motion of mouth. (b) Profiling the correlation of action unit's response between our result and the source, ReenactGAN exhibits a significantly higher correlation w.r.t. the source video.

To compare with Face2Face[35], we first collect its demo video[‡] as input, then reenact it to Trump. The result is compared with [35]'s result on Trump. We apply our facial AU detectors on the input video and two resulting videos, respectively from Face2Face and ReenactGAN.

In Figure 8(a) we show the response of three AUs over the time. Comparing the first two rows, the proposed method exhibits more synchronized response with the source input and yields higher AU scores. We also compute three typical AUs response correlations between the input and each output in Figure 8(b). ReenactGAN records significantly higher correlations against the baseline. The full comparison will be presented in the supplementary material.

### 4.4 Ablation Study on Boundary Transformer

In Sec. 3.2 we mentioned that by merely compose the encoder and a target-specific decoder, $\psi_T \cdot \phi$, one can already roughly reenact a source face to the target, as long as their face shapes adhere. Figure 9(a) provide a more detailed qualitative comparisons on the results with and without the transformer. Without the transformer, decoder sometimes generates blurred (row 2,4) results or images with severe textural artifacts (row 3).

In another experiment, we study the role of each term in the loss function (Eqn. (2)). Figure 9(b) shows the adherence of facial boundary given the source, when we sequentially add a loss term to the transformer loss. By adding shape constraint, the transformed poses track the source's pose better than applying the GAN loss alone. By adding the cycle loss, we can further refine the result. A careful inspection of the last two rows reveals that the cycle loss helps to improve the generation of eyes and mouth regions.

---

[‡]https://www.youtube.com/watch?v=ohmajJTcpNk

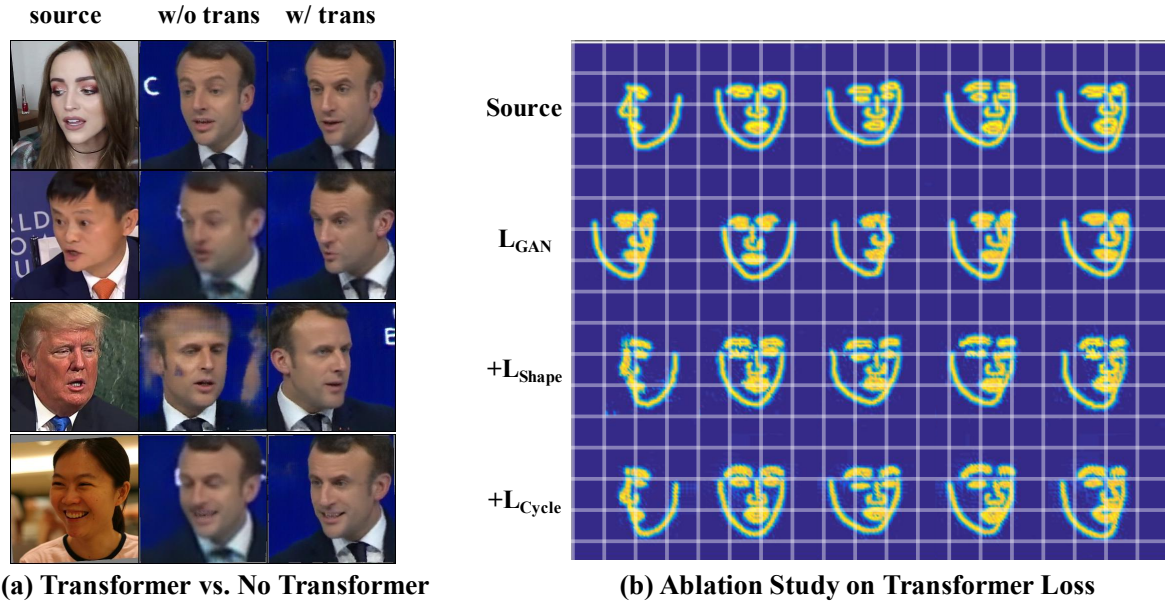

**Fig. 9. Ablation study on boundary transformer:** (a) Comparing the results of ReenactGAN with and without the transformer. (b) The effect of adding different loss terms to the transformer loss in Eqn. (2).

## 5   Conclusion and Discussion

We have presented a learning-based method, ReenactGAN, for photo-realistic face reenactment. Our method is novel in that we take a radically different approach to transfer source person to the target. Specifically, we bridge the source and target domains by introducing a boundary latent space. Our extensive qualitative and quantitative experiments showed that this boundary space is effective to reenact faces with accurate expressions and good quality. The many-to-one transformer also effectively extends the capability of the decoder. The trained model can comfortably handle faces from unseen people. Some improvements are feasible and obvious: 1) We can introduce facial component discriminators to enhance the generation on each facial part. 2) We may compress multiple target's decoders into one network, which is more efficient. 3) It will be interesting to investigate learning-based reenactment between faces of human and non-human (animal or cartoon characters).

**Acknowledgement.** We would like to thank Kwan-Yee Lin for insightful discussion, and Tong Li, Yue He and Lichen Zhou for their exceptional support. This work is supported by SenseTime Research.